\documentclass[acmlarge, nonacm]{acmart}
\usepackage[english]{babel}
\usepackage{comment}
\usepackage{amsmath,amsfonts}
\usepackage{algorithmic}
\usepackage{graphicx}
\usepackage{algorithm}
\usepackage{hyperref}
\hypersetup{hidelinks}
\usepackage{textcomp}
\usepackage{booktabs, multirow, makecell} 
\newcolumntype{P}[1]{>{\raggedright\arraybackslash\noindent}p{#1}} 

\AtBeginDocument{%
  \providecommand\BibTeX{{%
    \normalfont B\kern-0.5em{\scshape i\kern-0.25em b}\kern-0.8em\TeX}}}

\setcopyright{acmcopyright}
\copyrightyear{2018}
\acmYear{2018}
\acmDOI{XXXXXXX.XXXXXXX}

\acmConference[Conference acronym 'XX]{Make sure to enter the correct
  conference title from your rights confirmation emai}{June 03--05,
  2018}{Woodstock, NY}
\acmPrice{15.00}
\acmISBN{978-1-4503-XXXX-X/18/06}

\begin{document}

\title{Scaling Representation Learning from Ubiquitous ECG with State-Space Models}

\author{Kleanthis Avramidis}
\affiliation{%
  \institution{University of Southern California}
  \city{Los Angeles, CA}
  \country{USA}
}

\author{Dominika Kunc}
\affiliation{%
 \institution{Wrocław University of Science and Technology}
 \city{Wrocław}
 \country{Poland}}

 \author{Bartosz Perz}
\affiliation{%
 \institution{Wrocław University of Science and Technology}
 \city{Wrocław}
 \country{Poland}}

\author{Kranti Adsul}
\affiliation{%
  \institution{University of Southern California}
  \city{Los Angeles, CA}
  \country{USA}
}

\author{Tiantian Feng}
\affiliation{%
  \institution{University of Southern California}
  \city{Los Angeles, CA}
  \country{USA}
}

\author{Przemysław~Kazienko}
\affiliation{%
 \institution{Wrocław University of Science and Technology}
 \city{Wrocław}
 \country{Poland}}

\author{Stanisław Saganowski}
\affiliation{%
 \institution{Wrocław University of Science and Technology}
 \city{Wrocław}
 \country{Poland}}

\author{Shrikanth Narayanan}
\affiliation{%
  \institution{University of Southern California}
  \city{Los Angeles, CA}
  \country{USA}
}

\renewcommand{\shortauthors}{Avramidis et al. (2023)}

\begin{abstract}
Ubiquitous sensing from wearable devices in the wild holds promise for enhancing human well-being, from diagnosing clinical conditions and measuring stress to building adaptive health promoting scaffolds. But the large volumes of data therein across heterogeneous contexts pose challenges for conventional supervised learning approaches. Representation Learning from biological signals is an emerging realm catalyzed by the recent advances in computational modeling and the abundance of publicly shared databases. The electrocardiogram (ECG) is the primary researched modality in this context, with applications in health monitoring, stress and affect estimation. Yet, most studies are limited by small-scale controlled data collection and over-parameterized architecture choices. We introduce \textbf{WildECG}, a pre-trained state-space model for representation learning from ECG signals. We train this model in a self-supervised manner with 275,000 10s ECG recordings collected in the wild and evaluate it on a range of downstream tasks. The proposed model is a robust backbone for ECG analysis, providing competitive performance on most of the tasks considered, while demonstrating efficacy in low-resource regimes. The code and pre-trained weights are shared publicly at \href{https://github.com/klean2050/tiles\_ecg\_model}{github.com/klean2050/tiles\_ecg\_model}.

\end{abstract}


\maketitle

\section{Introduction}
\label{introduction}

Artificial Intelligence (AI) has made significant inroads into human-centered signal modeling, notably in the realms of behavioral analysis \cite{Bone2017SignalProcessingandMachine} and health \cite{yu2018artificial}. This progress benefits primarily from the algorithmic development of deep learning models and the substantial effort in curating publicly-shared datasets \cite{lecun2015deep}. The rapid advances of deep learning in various application domains, such as computer vision (CV), speech, and natural language processing (NLP) are critically dependent on the availability of large datasets, allowing for designing and training large-scale neural networks. Within the medical domain of biosignal analysis, supervised learning algorithms have been employed to improve diagnostic performance and accelerate biomarker detection in many areas, including dermatology \cite{esteva2017dermatologist}, ophthalmology \cite{dai2021deep}, as well as in psychology, physical health and well-being~\cite{rajpurkar2022ai, feng2021multimodal}\footnote{\textbf{ Pre-print, currently under review}. POC: avramidi@usc.edu}.

\begin{figure}
    \centering
    \includegraphics[scale=0.33]{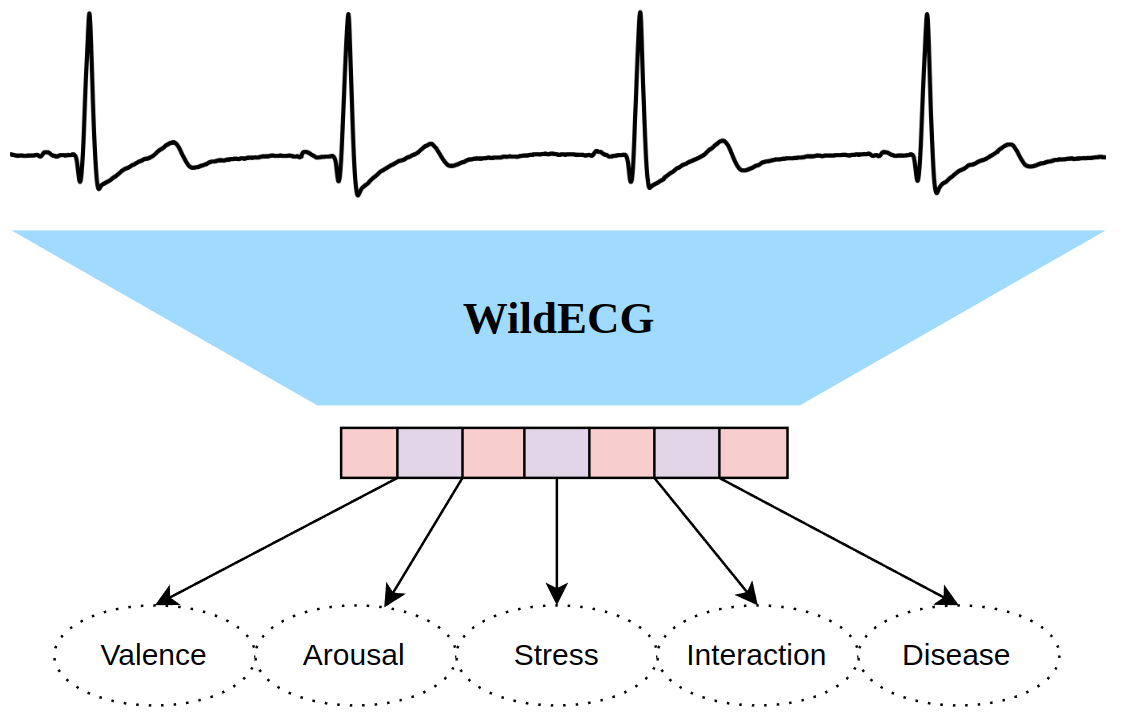}
    \caption{Our proposed model extracts vector representations from input, single-lead ECG signals, and can be used both as a backbone encoder and feature extractor across multiple different sensing tasks.}
    \label{fig:intro}
    \vspace{-0.5cm}
\end{figure}

Driven by successful applications in multiple fields within health and well-being, AI technologies are increasingly demanded in ubiquitous sensing and modeling of human states in everyday settings, including home and workplace~\cite{Thatte2012KNOWME:AnEnergyEfficientMultimodal, sano2013stress, saeb2015mobile, LHommedieu2019LessonsLearned:Recommendationsfor, saganowski2022bringing, booth2019toward}. A significant portion of research and understanding of human biosignals has focused on modeling physiological responses to external stimuli and from constrained interaction environments. These approaches typically consider signals derived from cardiac activity, respiration patterns, body temperature, electrodermal activity and even neural (brain) activity~\cite{rim2020deep}. Cardiac activity, particularly the electrocardiogram (ECG), has been a prominent modality choice due to its well-recognized signal patterns and clinically-validated significance \cite{minchole2019artificial,hannun2019cardiologist}. While it requires low-cost recording equipment, ECG offers enormous diagnostic potential and hence has seen increasing research efforts, notably through the creation of shared databases and novel, data-driven modeling approaches.

However, the transition from monitoring in clinical settings to sensing in the wild introduces novel challenges toward more generalized capabilities for understanding cardiac activity across diverse living contexts. Unlike text or image domains where abundant large datasets have enabled large (foundation) self-supervised models~\cite{gpt3}, the state of the art in biosignal models lags behind. One practical issue related to data acquisition is the need for long-term recording capabilities and the resulting cost of obtaining biological signals from ecologically valid contexts~\cite{booth2019multimodal}. This has challenged the creation of datasets with high-quality recordings from a large number of participants and diverse backgrounds. Furthermore, the intricacy of extracting meaningful insights from vast and heterogeneous ECG data, with sparse or no accompanying (meta) information, demands methods that can learn and infer from these data in self-supervised ways. Such methods are also motivated by the need to address inherent biases and subject heterogeneity in bio-behavioral responses that hinder model performance and reliability. Another critical challenge involves engineering models that can adeptly capture the structure and temporal dependencies of ECG without the need of scaling to large and overparameterized models~\cite{resnet} that increase the risk of overfitting and are not practical for mobile deployment. This involves striking a delicate balance between model complexity and efficiency. Given the multifaceted nature of these challenges, there is an imminent need for developing methodologies toward practical, robust, and reliable solutions in the realm of ubiquitous ECG analysis.

\section{Contributions}
\label{contributions}

In this study, toward addressing those challenges, we propose a framework to train a model on large-scale public data to extract general-purpose vector representations for the ECG signal. Our contributions can be summarized as follows:

\begin{itemize}
    \item Our model, called \textbf{WildECG}, is trained on TILES~\cite{mundnich2020tiles2018}, one of the largest publicly available biosignal data collections in the wild, manifesting a wide range of variability and subject heterogeneity for better generalization.
    \item To minimize the impact of noisy and biased data annotations, WildECG is trained in a self-supervised manner to identify distortions automatically induced on ECG samples at training time. This enables us to test performance on a variety of downstream tasks related to ECG.
    \item Our model incorporates a lightweight architecture based on state-space models that are efficient in modeling sequences with long temporal dependencies. By using a small number of parameters, WildECG further reduces the risk of overfitting and is suitable for deployment on the edge. This is particularly important for user privacy and data security by handling all computations locally.
\end{itemize}

In sum, the proposed framework offers an efficient way for extracting robust ECG representations that perform competitively across multiple downstream tasks including human interaction modeling, affect recognition, and disease prediction. WildECG outperforms multiple architectures and training algorithms, while retaining discriminative information in low-resource settings and during minimal fine-tuning.

\section{Background}
\label{background}

\subsection{Electrocardiography}
\label{back-ecg}

Electrocardiography is a non-invasive technique for recording the electrical activity of the heart. The resulting signal, called the electrocardiogram (ECG), provides information about the functioning and structure of the heart, including the timing and regularity of its rhythm, significant underlying conditions or abnormalities, along with psychological states such as stress or emotional arousal. ECG has a characteristic structure that consists of specifically documented signatures: the P wave, the R peak and broadly the QRS complex, and the T wave, each of which corresponds to a distinct phase of the cardiac cycle. In addition to these waves, one can also identify intervals on the ECG that hold important information, e.g., PR and QT intervals, as well as the RR interval which is used to calculate heart rate and heart rate variability \cite{valstar2016avec, koldijk2014swell, nashiro2023increasing, nashiro2023effects}.

ECG is acquired through electrodes that are placed on the surface of the skin, usually at the chest or the wrists. The most common sensor configurations include 12-lead, 3-lead, or single-lead placements. The latter configuration, applied through wearable straps or wristbands, is a practical choice for measurements made in naturalistic settings, due to the ease of placement and minimal interference with the subject. To facilitate applications in both clinical and naturalistic domains, we restrict our study to single-lead ECG data.

\subsection{Self-Supervised Learning}
\label{back-ssl}

Self-supervised learning (SSL) is an emerging machine learning paradigm that provides an effective way to learn meaningful data representation without the need to acquire explicit labels. In contrast to supervised learning, which relies on labeled data, SSL leverages the intrinsic structure and relationships within the data to create pseudo-labels or tasks to learn from. As such, it holds several advantages over conventional supervised approaches for our task, as it avoids the need for reliable annotations in large quantities, which would also constrain the scope of the model.

Most researchers distinguish two main types of SSL frameworks: (1) generative and (2) contrastive \cite{mohamed2022self, liu2021self, liu2022graph, yu2023self, wang2022self, schiappa2023self}. The generative models (e.g., autoencoders) learn representations by reconstructing or generating the original input data using masked or corrupted data as input, which defines their pretext task. Contrastive methods, on the other hand, train a model by contrasting the representations of semantically same data (e.g., two augmented views, positive samples) to other distant data (negative samples). Additional variants of SSL have also been proposed in the literature, including predictive \cite{mohamed2022self, yu2023self, wang2022self}, property-based \cite{liu2022graph} or pretext learning \cite{schiappa2023self} objectives, as well hybrid \cite{liu2022graph, yu2023self} or cross-modal \cite{schiappa2023self, liu2021self} types.

Several promising approaches to SSL have been implemented, primarily in natural language processing (NLP)~\cite{bert,wav2vec} and computer vision~\cite{jigsaw,chen2020simple}. In the context of time series data, SSL has been used to learn representations for various tasks such as anomaly detection~\cite{sarkar2020self}, frequency coupling~\cite{zhang2022self}, and masking~\cite{bendr}. Self-supervised learning of biosignals and ECG has already shown promising results in health applications and behavioral analysis. We include a comprehensive review of related studies in Appendix~\ref{related-ssl}. In this work, we design a custom data augmentation approach to automatically identify distortions induced on ECG samples, as a strategy to leverage the inherent structure and temporal dependencies that characterize the ECG signal.

\subsection{State-Space Models}
\label{back-ssm}

State-space models (SSM) are a recently introduced category of deep neural networks~\cite{gu2021combining} that were proposed to efficiently model long-term sequences, i.e., signals with either long duration and/or high sampling rate. Hence, the ECG modality constitutes a promising candidate for adopting state-space model architecture.  SSMs draw intuition from both convolutional and recurrent network architectures. The continuous-time SSM converts a 1-D input signal $u(t)$ into a latent state $x(t)$ before projecting it onto a 1-D output $y(t)$:
\begin{equation}\label{eq:SSM}
\begin{split}
x^{\prime}(t) = A x(t)+B u(t) \\
y(t) = C x(t)+D u(t)
\end{split}
\end{equation}
For discrete-time sequences that are sampled at a step $\Delta$, Eq.~\ref{eq:SSM} can be mapped to the recurrence shown in Eq.~\ref{eq:recurrence}, using the bilinear method \cite{tustin1947method} to convert $A$ into an approximation $\bar{A}$:
\begin{equation}\label{eq:recurrence}
\begin{split}
x_k = \bar{A} x_{k-1}+\bar{B} u_k \quad y_k=C x_k+D u_k \\
\bar{A} = (I-\Delta / 2 \cdot A)^{-1}(I+\Delta / 2 \cdot A) \\
\bar{B} = (I-\Delta / 2 \cdot A)^{-1}\Delta B
\end{split}
\end{equation}
Here $D=0$~\cite{s4}. Eq.~\ref{eq:recurrence} is a sequence-to-sequence map and the recurrence allows the discrete SSM to be computed like a recurrent network with hidden state $\bar{A}$. Eq.~\ref{eq:recurrence} is also equivalent to a discrete convolution with kernel $\bar{K}$~\cite{s4}:
\begin{equation}\label{eq:kernel}
\bar{K}=\left(C \bar{B}, C \overline{A B}, C \bar{A}^2 \bar{B}, \ldots\right), \quad y=\bar{K} * u
\end{equation}
Thus, SSMs can be viewed as special cases of convolutional and recurrent layers, inheriting their learning efficiency. Gu et al.~\cite{s4} also contributed an efficient way of evaluating $\bar{K}$.

The \textit{Structured State Space for Sequence Modeling} (S4) architecture was proposed in \cite{s4} to model sequences more efficiently than standard SSMs, also showing the capacity to capture long-range temporal dependencies. S4 is a particular instantiation of the SSM, where matrix $A$ is parameterized as a diagonal plus low-rank (DPLR) that allows faster repeated computations. To capture long-range dependencies, S4 initializes $A$ as HiPPO~\cite{gu2020hippo}, so that the state $x_k$ can memorize the history of the input $u_k$. At the same time, HiPPO preserves the DPLR form, as shown in \cite{s4}. Hence, the core S4 module is a linear, 1-D sequence mapping, however it handles high-dimensional features by defining independent copies of itself, and then mixing features with a position-wise linear layer. Nonlinear
activations and dropouts in-between these layers provide the non-linearity of the whole architecture.


\section{Method}
\label{method}

\subsection{Pre-Processing}
\label{method-preproc}

We adopt a universal approach in processing all ECG data used in pre-training and fine-tuning sessions. The following steps aim to alleviate the impact of discrepancies in different data collection processes, such as the performed task, sampling rate, equipment noise, subject-specific and other artifacts that induce different spatiotemporal properties. First, ECG signals are downsampled to 100 Hz and smoothed with a moving average kernel to remove powerline interference \cite{Makowski2021neurokit}. The specific sampling frequency provides a balance between preserving relevant information and reducing computational requirements. The majority of ECG datasets are recorded at 100 Hz or higher, and it has been reported~\cite{kwon2018electrocardiogram, strodthoff2020deep} that downsampling to 100 Hz does not compromise model performance. Next, we apply a high-pass Butterworth filter at 0.5 Hz. Finally, we perform subject-wise, z-score normalization. The signals are then segmented into non-overlapping windows of 10 second duration. During pre-training, where each sample is 15 seconds, 10 s samples are randomly extracted during training.

\subsection{Signal Transformations}
\label{method-trans}

We base our proxy tasks for pre-training on evaluating various signal transforms applied to ECG samples. To this end, we have implemented a Python module of ECG-tailored transformations that we share publicly\footnote{\href{https://github.com/klean2050/ecg-augmentations}{https://github.com/klean2050/ecg-augmentations}}. The ECG-augmentations library \cite{avramidis_kleanthis_2023_8053043} currently includes versatile transforms of multi-lead ECG signals. Implemented augmentations include:

\begin{itemize}
    \item \textbf{Masking}: We currently support random masking or masking of PR and QRS intervals, whereas the user can also specify the ratio of intervals to be masked. Detection of R peaks is done using NeuroKit2 \cite{Makowski2021neurokit}.
    \item \textbf{Cropping}: Random ($r$) cropping of an ECG sub-sequence given the desired length $\lambda$: $s' = s[r: r+\lambda]$
    \item \textbf{Noise}: We support both additive white noise and random wander, with adjustable signal-to-noise ratio.
    \item \textbf{Permutation}: Each ECG signal is divided into $m\leq10$ segments, which are shuffled, i.e., by randomly perturbing their temporal order. Each segment has a set minimum length of 10\% the total signal length.
    \item \textbf{Time Warping}: Randomly selected segments of the original ECG are stretched or squeezed along the temporal axis, through interpolation. The output signal is cropped or zero-padded when stretched or squeezed, respectively.
    \item \textbf{Scaling}: The ECG magnitude is multiplied by a random scalar $0<\alpha<5$: $s'[n] = -\alpha s[n], n=0, ..., N$
    \item \textbf{Inverting}: Implemented by negating the input signal along the temporal axis: $s'[n] = s[N-n], n=0, ..., N$
    \item \textbf{Reversing}: Simply implemented by scaling the input signal using $\alpha = -1$: $s'[n] = -s[n], n=0, ..., N$
\end{itemize}

\subsection{Pre-Training Objective}
\label{method-objective}

Most SSL studies apply either masked sample reconstruction or contrastive learning objectives to pre-train their respective models. Since there is no established training algorithm for physiological signals, WildECG considers elements from both SSL approaches. Our objective aims to identify which signal transformations are applied to a sample ECG, where each signal is augmented randomly using at most four out of all the available transforms. Each transform is selected based on a set probability, so it is possible that some samples are input without any augmentation. We formulate this task as a multi-label classification task with nine classes (eight possible transformations plus the original signal).

This task draws from both predominant SSL approaches. Our first motivation comes from masked reconstruction objectives by including masking augmentations in our pre-training framework. These include both random masking of signal patches and masking of specific ECG intervals. Second, we follow the contrastive learning paradigm in the sense that we evaluate the impact of induced augmentations in the data. However, we intentionally choose to predict applied transformations over the conventional contrastive approach, in which we would identify the similarity of two distorted samples of the same input. The reason is that this objective would focus on invariant ECG features that are primarily subject-dependent. On the other hand, identifying distortions is intuitive for our scope, since the model focuses on ECG abnormalities that could potentially hold diagnostic information.

\subsection{Model Architecture}

\begin{figure}
    \centering
    \includegraphics[scale=0.24]{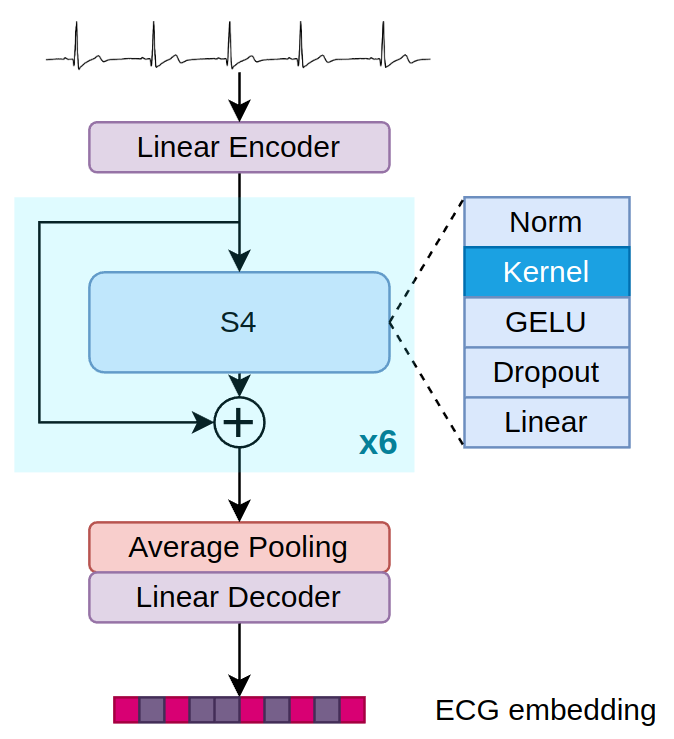}
    \caption{The architecture of the proposed ECG backbone model, following a simple version of the original S4~\cite{s4}. The model consists of six S4 blocks, connected through residual connections. Linear classifiers are attached on top.}
    \label{fig:architecture}
\end{figure}

Our proposed model inherits the S4 model as the backbone architecture, as it features critical elements that are desirable and intuitive in ECG analysis. As mentioned before, S4 has demonstrated promising performance in modeling long-range sequences with dependencies over thousands of timesteps, which is a current limitation of state-of-the-art models like the Transformer~\cite{zaheer2020big} architecture. ECG is a sequence of that type, with its sampling rate ranging from 100 to 1000 Hz. Also, S4 is implicitly a continuous-time model, making it well-suited to waveform signals. Indeed, prior work \cite{zhang2023effectively} has shown that variants of S4 provide excellent performance in classifying cardiovascular conditions based on controlled ECG data.

Here we employ a simplified version of the original S4, consisting of a linear encoder, six S4 blocks, and a linear decoder. Each block consists of a Layer Normalization module, the $\bar{K}$ estimation module, a GELU \cite{hendrycks2016gaussian} activation, Dropout and an output projector. The blocks are connected with residual connections, as shown in Figure~\ref{fig:architecture}. The input and output dimension is set to 256 and dropout layers of 20\% are applied. For the pre-training phase, we adjust a linear layer to the decoder output and replace it during fine-tuning.

\begin{table}
\centering
\small
\caption{Overview of the study's datasets. $\star$ denotes regression.}
  \label{tab:datasets}
  \begin{tabular}{@{}lccc@{}}
    \toprule
    \textbf{Dataset} & \textbf{ECG Setting} & \textbf{\# Subjects} & \textbf{\# Classes} \\
    \midrule
    TILES \cite{mundnich2020tiles2018} & 24-h monitoring & 200 & N/A  \\
    \midrule
    PTB-XL \cite{wagner2020ptb} & clinical acquisition & 18869 & 5  \\
    LUDB \cite{kalyakulina2020ludb} & clinical acquisition & 200 & 2  \\
    WESAD \cite{schmidt2018introducing} & activity engagement & 15 & 3  \\
    CASE \cite{sharma2019dataset} & video watching & 30 & $\star$  \\
    AVEC-16 \cite{valstar2016avec} & dyadic interaction & 27 & $\star$  \\
    SWELL-KW \cite{koldijk2014swell} & workplace stress & 25 & 2  \\
  \bottomrule
  \end{tabular}
\end{table}

\section{Experiments}
\label{experiments}

Below we share details about the datasets of this study (Table~\ref{tab:datasets}). Our experimentation covers 7 widely used ECG datasets and targets settings where the 1-lead ECG modality is prominent and the evaluation criteria are clearly defined. We thus omitted datasets such as DEAP~\cite{koelstra2011deap}, AMIGOS~\cite{miranda2018amigos}, or DREAMER~\cite{katsigiannis2017dreamer} as EEG-oriented, and also medical datasets that depend heavily on 12-lead ECG recordings.

\subsection{Pre-training: TILES Dataset}

Tracking Individual Performance with Sensors (TILES) \cite{mundnich2020tiles2018, yau2022tiles} is a research project that has collected multimodal data sets for the analysis of stress, task performance, behavior, and other factors pertaining to professionals in a high-stress workplace environment. Biological, environmental, and contextual data were collected from hospital nurses, staff, and medical residents both in the workplace and at home over a ten week period. Labels of human experiences, such as stress, anxiety, and affect, were collected using psychologically validated questionnaires which were administered at different times.

In the present study, we use the ECG data from the publicly available TILES 2018 dataset \cite{mundnich2020tiles2018} to pre-train a general-purpose ECG model. Each participant had their ECG recorded for 15 seconds every 5 minutes during their work hours, for a total of 10 weeks. There were 213 participants in total, 200 of whom had agreed to wear a bioshirt that enabled high quality ECG data collection, making the aggregate number of samples conducive to pre-train a large ECG representation model. Since ECG was collected in the wild, we apply a quality check on the available signals by attempting to identify RR intervals. All days for which the total detection rate is lower than 90\% are discarded, leading to approximately 275,000 ECG samples from 168 individuals.

\subsection{Fine-tuning: In-the-wild Sensing}
\label{data-active}

\subsubsection{AVEC-16 Multimodal Affect Recognition Sub-challenge}

The multimodal affect recognition sub-challenge (MASC) of AVEC-16 \cite{valstar2016avec} stems from the REmote COLlaborative and Affective interactions (RECOLA) dataset \cite{ringeval2013introducing}. RECOLA included continuous multimodal recordings during dyadic interactions via video conferences. 
The complete dataset contains audio, visual, and physiological information from 27 French-speaking participants. The single-channel ECG data used in this work were sampled at 250 Hz and subsequently filtered using a band-pass filter at 3-27 Hz. The labels are continuous ratings for arousal and valence at 40ms intervals, throughout the first five minutes of the complete recordings. 

\subsubsection{SWELL-KW}

This dataset \cite{koldijk2014swell} aimed at analyzing employees' emotional states and workplace stress under three scenarios: \textit{normal}, in which participants performed various office tasks for 45 minutes, \textit{time-pressure}, in which participants had only 30 minutes to complete the same tasks, and \textit{interruption}, in which they were also interrupted by emails and messages. ECG signals were collected from 25 participants using the TMSI MOBI device at a sampling rate of 2048 Hz. At the end of each scenario, participants were asked to report their valence, arousal, and also other states, such as stress.

\subsubsection{WESAD}

The dataset for WEarable Stress and Affect Detection (WESAD) \cite{schmidt2018introducing} contains ECG data from 15 participants. RespiBAN Professional sensors were used to collect ECG at a sampling rate of 700 Hz. The goal was to study four different affective states (neutral, stressed, amused, and meditated). First, 20 minutes of neutral condition data were collected, during which participants were asked to do normal activities. Then participants watched 11 funny video clips (amusement) and went through public speaking and arithmetic tasks (stress). Finally, they went through a guided meditation session of 7 minutes. Upon completion of each trial, labels for the affect states were collected using 9-scale PANAS.

\subsection{Fine-tuning: Passive Sensing}
\label{data-passive}

\subsubsection{PTB-XL}

The PTB-XL dataset \cite{wagner2020ptb} is a set of 21799 clinical 12-lead ECGs from 18869 patients of 10 second length. The raw waveform data were annotated by up to two cardiologists, who assigned potentially multiple ECG statements to each record. The waveform data underlying the PTB-XL ECG dataset were collected with devices from Schiller AG over the course of nearly seven years between October 1989 and June 1996. In total 71 different ECG statements conform to the SCP-ECG standard and cover diagnostic, form, and rhythm statements. The dataset is complemented by extensive metadata on demographics, infarction characteristics, diagnostic statements, and annotated signal properties.

\subsubsection{LUDB}

The Lobachevsky University Electrocardiography Database (LUDB) \cite{kalyakulina2020ludb} is an ECG dataset with annotated boundaries and peaks of P, T and QRS waves. It consists of 200 10-second ECG signals at 500 Hz, representing different morphologies, out of which we only use the first lead, to comply with our framework. The ECG records were collected from healthy volunteers and patients of the Nizhny Novgorod City Hospital during 2017–2018. The patients had various cardiovascular diseases while some of them had pacemakers. Cardiologists annotated each record with the corresponding diagnosis. For this study, we consider the task of identifying sinus rhythm against a super-set of different abnormalities.

\subsubsection{CASE}

The Continuously Annotated Signals of Emotion (CASE) dataset \cite{sharma2019dataset} contains data from 30 participants collected in laboratory conditions. During the experiment, participants watched a series of 8 video stimuli and continuously annotated their emotions in a two-dimensional arousal-valence space using a joystick interface developed by the researchers. Additionally, a two-minute long blue-screen video served as an in-between resting phase. The ECG data were collected at 1000 Hz using Thought Technology SA9306 sensors, and affect annotations were collected at 20 Hz.

\begin{table}
\centering
\small
\caption{Downstream performance on \textbf{AVEC-16} dataset (Eval Split).}
  \label{tab:results_avec}
  \begin{tabular}{@{}lccc@{}}
    \toprule
    \textbf{Model} & \textbf{Training} & \textbf{Arousal CCC} & \textbf{Valence CCC} \\
    \midrule
    Linear SVM \cite{valstar2016avec} & -- & 0.271 & 0.153 \\
    Linear SVM \cite{2018avec} & -- & 0.118 & 0.085
    \\
    \midrule
    Baseline S4 & full model & 0.328 & 0.162  \\
    WildECG (ours) & full model & \textbf{0.356} & \textbf{0.303}  \\
    WildECG (ours) & projector & 0.346 & 0.289  \\
  \bottomrule
  \end{tabular}
\end{table}

\begin{table*}
\centering
\small
\caption{Downstream performance on \textbf{WESAD} dataset (3-way activity). $^*$ Pre-training includes WESAD. Standard deviation among folds is included in parentheses.}
\label{tab:results_wesad}
  \begin{tabular}{@{}l@{\hskip 6pt}c@{\hskip 6pt}c@{\hskip 5pt}c@{\hskip 5pt}c@{\hskip 5pt}c@{}}
    \toprule
    \multicolumn{1}{@{}l}{{\textbf{Model}}} & \multicolumn{1}{c}{{\textbf{Training}}} & \multicolumn{2}{c}{\textbf{Mixed-subject}} & \multicolumn{2}{c@{}}{\textbf{Subject-agnostic}} \\
    
    \cmidrule(lr){3-4} \cmidrule(l){5-6} & &
    \textbf{Accuracy} & \textbf{F1-macro} & \textbf{Accuracy} & \textbf{F1-macro} \\
    \midrule
    AdaBoost \cite{schmidt2018introducing} & -- & -- & -- & 0.617 ( -- ) & 0.525 ( -- ) \\
    LDA \cite{schmidt2018introducing} & -- & -- & -- & \textbf{0.663} ( -- ) & 0.560 ( -- ) \\
    1D-CNN \cite{sarkar2020self} & projector $^*$ & 0.969 ( -- ) & 0.963 ( -- ) & -- & -- \\
    \midrule
    Baseline S4 & full model & 0.956 (0.031) & 0.955 (0.032) & 0.489 (0.089) & 0.410 (0.097) \\
    WildECG (ours) & full model & \textbf{0.978} (0.028) & \textbf{0.978} (0.028) & 
    0.644 (0.044) & \textbf{0.592} (0.058) \\
    WildECG (ours) & projector & 0.742 (0.044) & 0.721 (0.064) & 0.600 (0.089) & 0.524 (0.075) \\
  \bottomrule
  \end{tabular}
\end{table*}

\begin{table*}
\centering
\small
\caption{Downstream performance on \textbf{WESAD} dataset (stress vs normal -- subject-agnostic).}
  \label{tab:results_wesad_stress}
  \begin{tabular}{@{}l@{\hskip 8pt}c@{\hskip 6pt}c@{\hskip 5pt}c@{}}
    \toprule
    \textbf{Model} & \textbf{Training} & \textbf{Accuracy} & \textbf{F1-macro} \\
    \midrule
    LDA \cite{schmidt2018introducing} & -- & 0.854 & 0.813 \\
    2D-CNN \cite{liakopoulos2021cnn} & full model & 0.824 & 0.794 \\
    Transformer \cite{behinaein2021transformer} & full model & 0.804 & 0.697 \\
    ECGNet \cite{prajod2022generalizability} & full model & 0.908 & 0.857 \\
    SVM \cite{henry2023robustness} & -- & 0.811 & 0.818 \\
    \midrule
    Baseline S4 & full model & 0.900 & 0.899  \\
    WildECG (ours) & full model & \textbf{0.967} & \textbf{0.966} \\
    WildECG (ours) & projector & 0.900 & 0.891  \\
  \bottomrule
  \end{tabular}
\end{table*}

\begin{table*}
\centering
\small
\caption{Downstream performance on \textbf{SWELL-KW} dataset for the subject-agnostic setting (binary classification). \\ Standard deviation among folds is included in parentheses. $^*$ not only ECG}
  \label{tab:results_swell}
  \begin{tabular}{@{}lcc@{\hskip 4pt}cc@{\hskip 4pt}cc@{\hskip 4pt}c@{}}
    \toprule
    \multicolumn{1}{@{}l}{{\textbf{Model}}} & \multicolumn{1}{c}{{\textbf{Training}}} & \multicolumn{2}{c}{\textbf{Valence $>$ mean}} & \multicolumn{2}{c}{\textbf{Arousal $>$ mean}} & \multicolumn{2}{c@{}}{\textbf{Stress (N vs T/I)}} \\
    
    \cmidrule(lr){3-4} \cmidrule(lr){5-6} \cmidrule(l){7-8} & &
    \textbf{Accuracy} & \textbf{F1-macro} & \textbf{Accuracy} & \textbf{F1-macro} & \textbf{Accuracy} & \textbf{F1-macro} \\
    \midrule
    SVM $^*$ \cite{koldijk2016detecting} & -- & -- & -- & -- & -- & 0.641 ( -- ) & -- \\
    Transformer \cite{behinaein2021transformer} & full model & -- & -- & -- & -- & 0.581 ( -- ) & 0.588 ( -- ) \\
    Deep ECGNet \cite{prajod2022generalizability} & full model & -- & -- & -- & -- & \textbf{0.755} ( -- ) & 0.688 ( -- ) \\
    \midrule
    S4 baseline & full model & 0.598 (0.098) & 0.560 (0.105) & 0.743 (0.153) & \textbf{0.731} (0.148) & 0.680 (0.075) & 0.643 (0.107) \\
    WildECG (ours) & full model & \textbf{0.629} (0.050) & \textbf{0.623} (0.050) & \textbf{0.751} (0.064) & 0.704 (0.077) & 0.660 (0.080) & 0.637 (0.094) \\
    WildECG (ours) & projector & 0.607 (0.118) & 0.560 (0.157) & 0.731 (0.088) & 0.698 (0.089) & 0.740 (0.102) & \textbf{0.711} (0.127) \\
  \bottomrule
  \end{tabular}
\end{table*}

\subsection{Implementation Details}

We pre-train WildECG for 100 epochs on the TILES data using a batch size of 256 samples and an AdamW optimizer with a 0.001 learning rate. A linear layer is used to map the ECG embeddings to the transform classes. We checkpoint the resulting model of the last epoch and apply it to a set of downstream tasks to evaluate the learned representations. For each task, the respective ECG data are extracted and processed akin to the TILES data (see Section~\ref{method-preproc}), whereas the additive linear layer is replaced by a 2-layer MLP classifier that maps the pre-trained embeddings to the target space.

We evaluate each task with 5-fold cross-validation in primarily subject-agnostic settings. \textit{Subject-agnostic} refers to the setting where test splits do not contain samples from subjects of the training splits, whereas \textit{mixed-subject} denotes the opposite. We use a batch size of 256 samples in all experiments except LUDB (32 samples). The learning rate is tuned to each dataset separately, within $\{0.0001, 0.0005, 0.001\}$. All datasets are trained for a maximum of 200 epochs with early stopping based on validation loss. Model checkpoints are selected based on the highest F1-macro or CCC in cross-validation, and lowest validation loss for PTB-XL and AVEC-16, which have specified validation and test splits.


\section{Results}
\label{sec:results}

Below we present the results of the downstream evaluation of WildECG. Our primary objective is to highlight the performance of the proposed model when employed both as a backbone and as a feature extractor, compared to training supervised classifiers. Wherever possible, we compare our performance with available studies, and when no comparable studies were found, we benchmark the respective task.

\subsection{In-the-wild Sensing}

\vspace{1mm}
\noindent \textbf{AVEC-16:} Table~\ref{tab:results_avec} includes results for the AVEC-16 dataset, quantified using the Concordance Correlation Coefficient (CCC). S4 persistently outperforms the scores reported from all prior studies, which rely on knowledge-based ECG features and conventional classifier architectures. For arousal estimation, WildECG achieves a state-of-the-art CCC of 0.356 when fully fine-tuned and 0.346 when only the projector is trained. In both cases it outperforms the S4 variant that is trained from scratch. Similar results are obtained for valence, where our proposed model surpasses 0.3 CCC.

\vspace{1mm}
\noindent \textbf{WESAD:} Table~\ref{tab:results_wesad} presents detailed results for WESAD, evaluated in both mixed-subject and subject-agnostic settings. Here, the objective is to identify the type of activity the subject performs out of three scenarios: baseline, stress, and amusement. For the mixed-subject setting, we compare our performance with Sarkar and Etemad~\cite{sarkar2020self} where we observe marginal improvements of 1\% to 1.5\% in F1-macro. We should note that the results are close to absolute correct accuracy, which is attributed to the temporal correlation that each subject and each recording inherits. This is evident by observing the drop of 23-25 percentage points (pp) when we freeze the pre-trained encoder, as well as the drop of more than 30 pp that the subject-agnostic setting induces. Nonetheless, WildECG outperforms the literature in obtained F1-macro, reaching 59.2\%, with an accuracy of 64.4\%. In this case, the pre-training mechanism is critical, since the S4 baseline cannot reach an accuracy better than random chance.

In addition to evaluating the 3-way condition in WESAD, we also assess the binary task of stress versus the two other conditions and report related results in Table~\ref{tab:results_wesad_stress}. WildECG achieves very high accuracy, reaching 96.6\% F1-macro and outperforming all previous studies by a large margin. Both frozen and fully fine-tuned models outperform convolutional and transformer encoders by 11--17\% F1-macro despite having fewer parameters.

\begin{table*}
\centering
\footnotesize
\caption{Downstream performance on \textbf{CASE} dataset (regression task). Standard deviation is included in parentheses.}
\label{tab:results_case}
  \begin{tabular}{lccccccc}
  
    \toprule
    
    \multicolumn{1}{@{}l}{{\textbf{Model}}} & \multicolumn{1}{c}{{\textbf{Training}}} & \multicolumn{3}{c}{\textbf{Mixed-subject}} & \multicolumn{3}{c@{}}{\textbf{Subject-agnostic}} \\
    
    \cmidrule(lr){3-5} \cmidrule(l){6-8} & &
    \textbf{Arousal CCC} & \textbf{Valence CCC} & \textbf{Anxiety CCC} & \textbf{Arousal CCC} & \textbf{Valence CCC} & \textbf{Anxiety CCC} \\
    
    \midrule
    
    Baseline S4 & full model & 0.249 (0.121) & 0.231 (0.133) & 0.292 (0.168) & 0.198 (0.085) & 0.162 (0.066) & 0.310 (0.159) \\
    
    WildECG (ours) & full model & \textbf{0.391} (0.047) & \textbf{0.439} (0.081) & \textbf{0.565} (0.089) & \textbf{0.253} (0.064) & \textbf{0.351} (0.086) & \textbf{0.424} (0.049) \\
    
    WildECG (ours) & projector & 0.226 (0.098) & 0.114 (0.066) & 0.219 (0.112) & 0.202 (0.034) & 0.123 (0.053) & 0.205 (0.058) \\
    
    \bottomrule
    
  \end{tabular}
\end{table*}

\begin{table}
\centering
\small
\caption{Downstream performance on \textbf{PTB-XL} dataset (sup-diag task).}
  \label{tab:results_ptbxl}
  \begin{tabular}{@{}lccc@{}}
    \toprule
    \textbf{Model} & \textbf{Training} & \textbf{AUROC} & \textbf{F1-macro} \\
    \midrule
    12-lead LSTM \cite{strodthoff2020deep} & full model & 0.927 & -- \\
    12-lead Inception-1D \cite{strodthoff2020deep} & full model & 0.921 & -- \\
    12-lead Transformer \cite{zhang2023effectively} & full model & 0.887 & -- \\
    12-lead S4 \cite{zhang2023effectively} & full model & 0.931 & -- \\
    \midrule
    1-lead S4 baseline & full model & 0.832 & 0.457  \\
    WildECG (ours) & full model & \textbf{0.845} & \textbf{0.480}  \\
    WildECG (ours) & projector & 0.815 & 0.346  \\
  \bottomrule
  \end{tabular}
\end{table}

\begin{table}
\centering
\small
\caption{Downstream performance on \textbf{LUDB} dataset (binary task). Standard deviation among folds is included in parentheses. }
  \label{tab:results_ludb}
  \begin{tabular}{@{}lccc@{}}
    \toprule
    \textbf{Model} & \textbf{Training} & \textbf{Accuracy} & \textbf{F1-macro} \\
    \midrule
    1-lead S4 baseline & full model & 0.770 (0.040) & 0.618 (0.141)  \\
    WildECG (ours) & full model & \textbf{0.915} (0.064) & \textbf{0.894} (0.082)  \\
    WildECG (ours) & projector & 0.855 (0.062) & 0.792 (0.094)  \\
  \bottomrule
  \end{tabular}
\end{table}

\vspace{1mm}
\noindent \textbf{SWELL-KW:}
Finally, we present results on SWELL-KW in Table~\ref{tab:results_swell}. Here we only share results for the subject-agnostic setting, since the mixed-subject one quickly overfits to perfect 100\% accuracy, with similar results shown in \cite{sarkar2020self}. For SWELL-KW, we evaluate three different binary cases: valence and arousal estimation, both binarized at the mean of the obtained values, and stress, as indicated again by the activity performed by the subject. For this task, $N$ refers to the normal condition whereas $T$ and $I$ represent the stress conditions. WildECG performs on par with the S4 trained from scratch, with smaller variations among folds in arousal. However, most studies evaluate their methods on the latter task of stress estimation, achieving more than 75\% accuracy and close to 69\% F1-macro. Our models provide competitive performance with these studies, with the pre-trained model reaching state-of-the-art 71.1\% F1-macro.

\subsection{Passive Sensing}

Here we report our performance scores for the datasets described in Section~\ref{data-passive}. We begin with the CASE dataset which incorporates target variables of affect. Table~\ref{tab:results_case} contains results for regression on arousal~(A), valence~(V), and anxiety~(N) levels, where anxiety is defined as ${N = A(1-V)}$, as proposed in \cite{henry2023robustness}. The prediction results show strong performance for WildECG, which outperforms both the S4 baseline and the frozen variant with a substantial margin in both subject-agnostic and mixed-subject settings. While the baseline shows higher performance for arousal, WildECG is better on valence, gaining 0.21 CCC from baseline S4. This difference is also reflected and magnified in the anxiety measures. We further observe that even though WildECG embeddings on their own offer limited performance improvements, our method leads to more consistent predictions in all tasks, with substantially lower variance than the baseline. To the best of our knowledge, no previous study provides continuous estimates of affect variables from ECG signals in CASE.

Although our system is not trained on clinical settings and data, we evaluate its performance as an out-of-distribution task on PTB-XL, one of the largest clinical ECG testbeds that is publicly available. Unfortunately, there is no extensive research assessing single-lead ECG systems for disease diagnosis. Moreover, many studies in cardiology report prominent disease biomarkers on several leads of a clinical ECG recording \cite{green2007best,reyna2021will}. With that premise, we compare our single-lead results with 12-lead ECG systems (Table~\ref{tab:results_ptbxl}). An S4 architecture similar to ours recently reported state-of-the-art performance \cite{zhang2023effectively} for 12-lead PTB-XL. Here we effectively benchmark the performance drop of 1-lead S4 to 10\% AUROC, reaching 83.2\%. We also demonstrate that when pre-trained on TILES, our model can improve upon this baseline by about 1.3 pp in AUROC and 2.3 pp in F1-macro. We also highlight that, possibly due to the scale of the dataset, the frozen model substantially under-performs in this 5-way classification task, with a 13\% drop in F1-macro. Further, in Table~\ref{tab:results_ludb}, we evaluate our model in LUDB, a much smaller medical dataset of various cardiac conditions. Despite the low-resource setting, WildECG distinguishes between healthy and non-healthy recordings, with a mean accuracy of 91.5\% and F1-macro of about 90\%, demonstrating data efficiency.

\begin{figure}
    \includegraphics[scale=0.7]{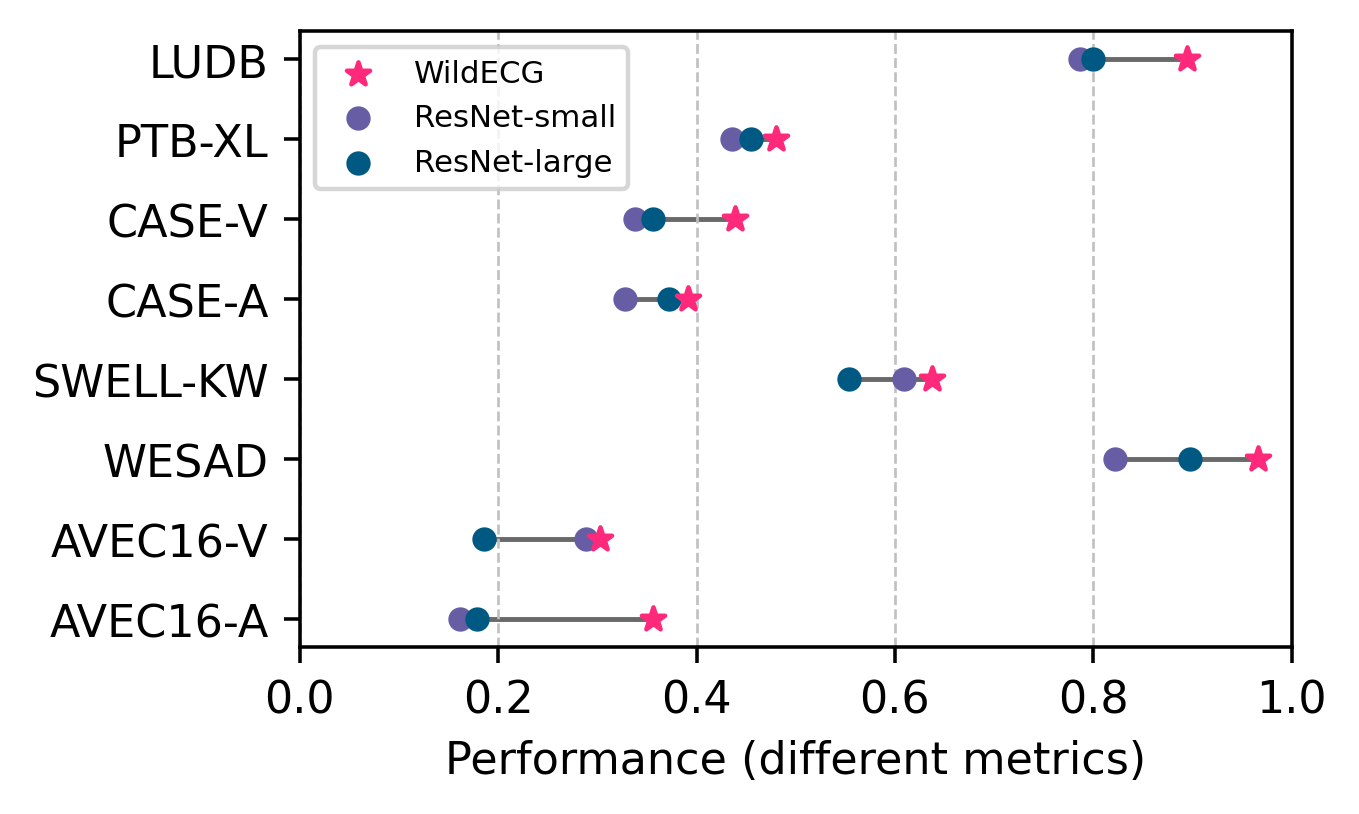}
    \includegraphics[scale=0.7]{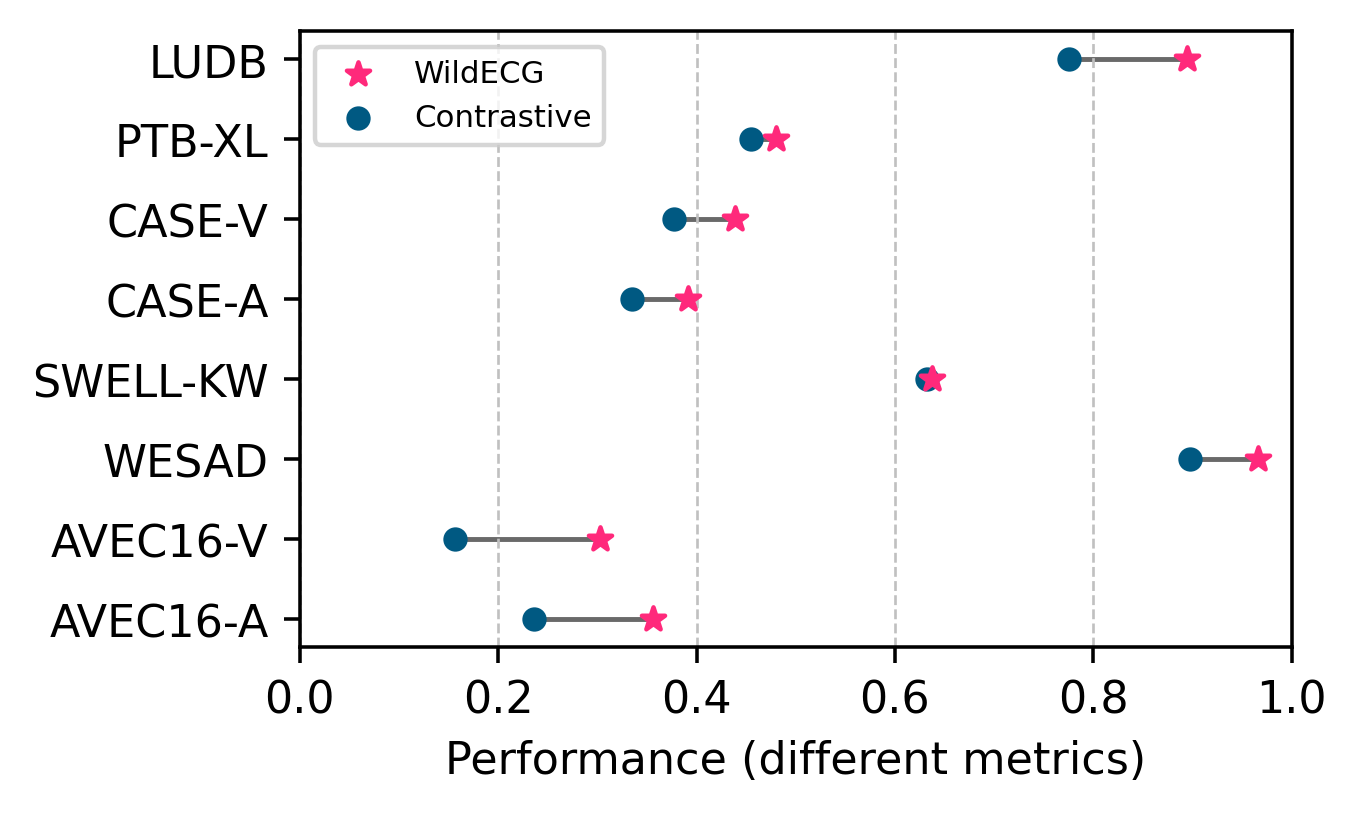}
    \caption{Left: Backbone architecture comparison. Right: Pre-training algorithm comparison. V denotes valence, A denotes arousal. WESAD and SWELL refer to the binary stress condition. All classification tasks are measured with F1-macro, all regression tasks with CCC.}
    \label{fig:ablations}
\end{figure}

\section{Discussion}

\subsection{Pre-Training Settings}

Our proposed framework incorporates several design parameters that contribute to positive experimental performance. In this section, we conduct a close inspection of each of these design elements by probing and comparing alternative approaches in the literature. Specifically, we compare WildECG to a network that uses a 1D ResNet~\cite{resnet} backbone, in order to assess the additive value of the selected architecture. ResNets have shown to be superior to other modeling approaches in a recent review on ECG signals~\cite{nonaka2021depth}. We create two versions, a ResNet-large of 14.7M parameters that includes 10 residual blocks and an input size of 64 filters, and a ResNet-small of 923K parameters by reducing filter sizes to $1/4$ of ResNet-large. We note that WildECG holds 313K parameters in total. Both networks are trained like WildECG and the obtained results for the fully trained variants are in Figure~\ref{fig:ablations} (up). We observe that our proposed S4 model clearly outperforms both ResNet variants except for SWELL-KW where the accuracy is similar. On the other hand, increasing the parameters of the ResNet model provides limited benefits to performance boost.

We further compare the ECG representations of our pre-training algorithm with those obtained using the standard contrastive learning approach~\cite{chen2020simple}. For this purpose, we train an identical to WildECG network with the alternative objective. The results on the same downstream tasks are shown in Figure~\ref{fig:ablations} (down). The chosen objective shows clear advantages over the contrastive one, as seen by the persistently improved performance in most evaluation cases. These results support our choice of pre-training objective as more suitable as well as more intuitive to ECG signal analysis. In future work, it is worth investigating tuning the baseline contrastive objective in order to reduce subject bias in the representations, for which Cheng et al.~\cite{cheng2020subject} have provided proof of concept by configuring subject-specific negative pairing.

\subsection{Low-Resource Scenarios}

\begin{figure}
    \centering
    \includegraphics[scale=0.49]{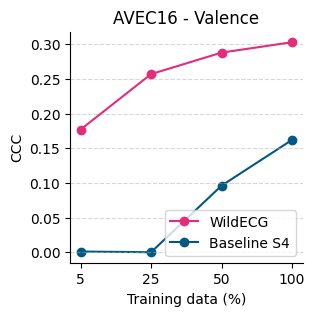}
    \includegraphics[scale=0.49]{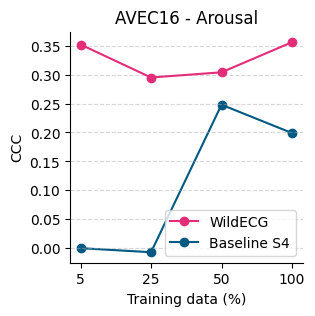}
    \includegraphics[scale=0.49]{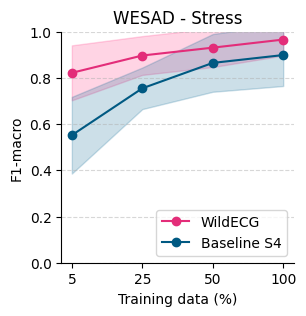}
    \includegraphics[scale=0.49]{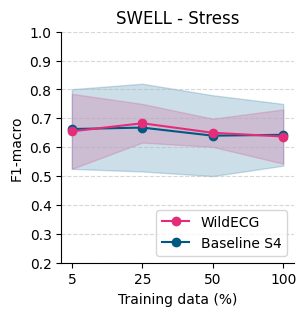}
    \caption{Low-resource model performance for AVEC-16, WESAD, and SWELL-KW: \textbf{active sensing}.}
    \label{fig:lowres_active}
\end{figure}

\begin{figure}
    \centering
    \includegraphics[scale=0.49]{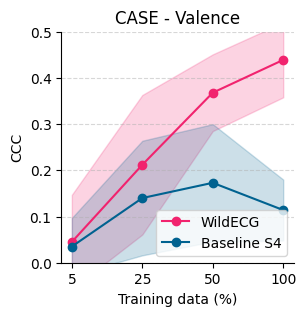}
    \includegraphics[scale=0.49]{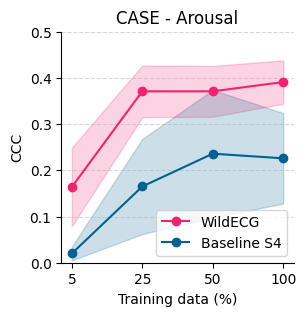}
    \includegraphics[scale=0.49]{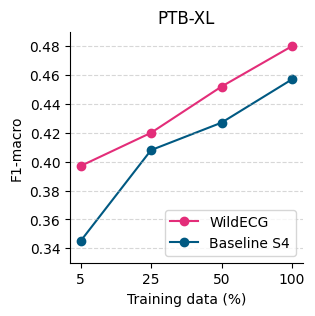}
    \includegraphics[scale=0.49]{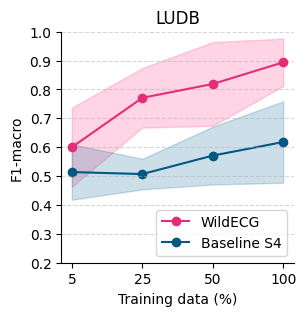}
    \caption{Low-resource model performance for CASE, PTB-XL, and LUDB: \textbf{passive sensing}.}
    \label{fig:lowres_passive}
\end{figure}

Thus far we have quantified our model's superiority against other architecture choices and across diverse applications. Herein we evaluate WildECG in low-resource settings, where we randomly restrict the number of training samples in each of the downstream tasks. Figures~\ref{fig:lowres_active} and~\ref{fig:lowres_passive} contain a respective graph for each dataset, with the horizontal axis denoting the percentage of the training samples that were actually used. In this section, we include the same subset of labels that we used previously. We observe that WildECG achieves small or even negligible performance loss in most cases compared to the baseline S4 network. For AVEC16, we are able to retain state-of-the-art performance even with 5\% of the training data, while the baseline fails to converge. Similarly, model pre-training alleviates the performance drop for WESAD and PTB-XL, whereas for LUDB the baseline never reaches performance substantially above chance. As mentioned earlier, SWELL-KW is associated with unnoticeable performance differences between WildECG and the S4 baseline, while our model does exhibit smaller variance in its predictions.

\subsection{Method Interpretability}

Although the proposed model shows state-of-the-art performance across different tasks, it is still unclear how the ECG is processed within the model and whether the derived representations have any feature-wise or semantic correspondence, which is critical for adoption in clinical practice. The field of explainable AI~\cite{amann2020explainability} has already made substantial progress in deciphering what deep learning models can learn. For the scope of this study, we provide post-hoc interpretability measures by mapping the high-dimensional ECG embeddings to a 2D space and analyzing between-sample distances. To reduce the dimensionality we use the t-SNE algorithm~\cite{van2008visualizing}. Since the TILES dataset incorporates a relatively large number of samples, we randomly select 10\% of its data (every 10th sample of each subject) to avoid over-plotting.

\begin{figure}
    \centering
    \includegraphics[scale=0.77]{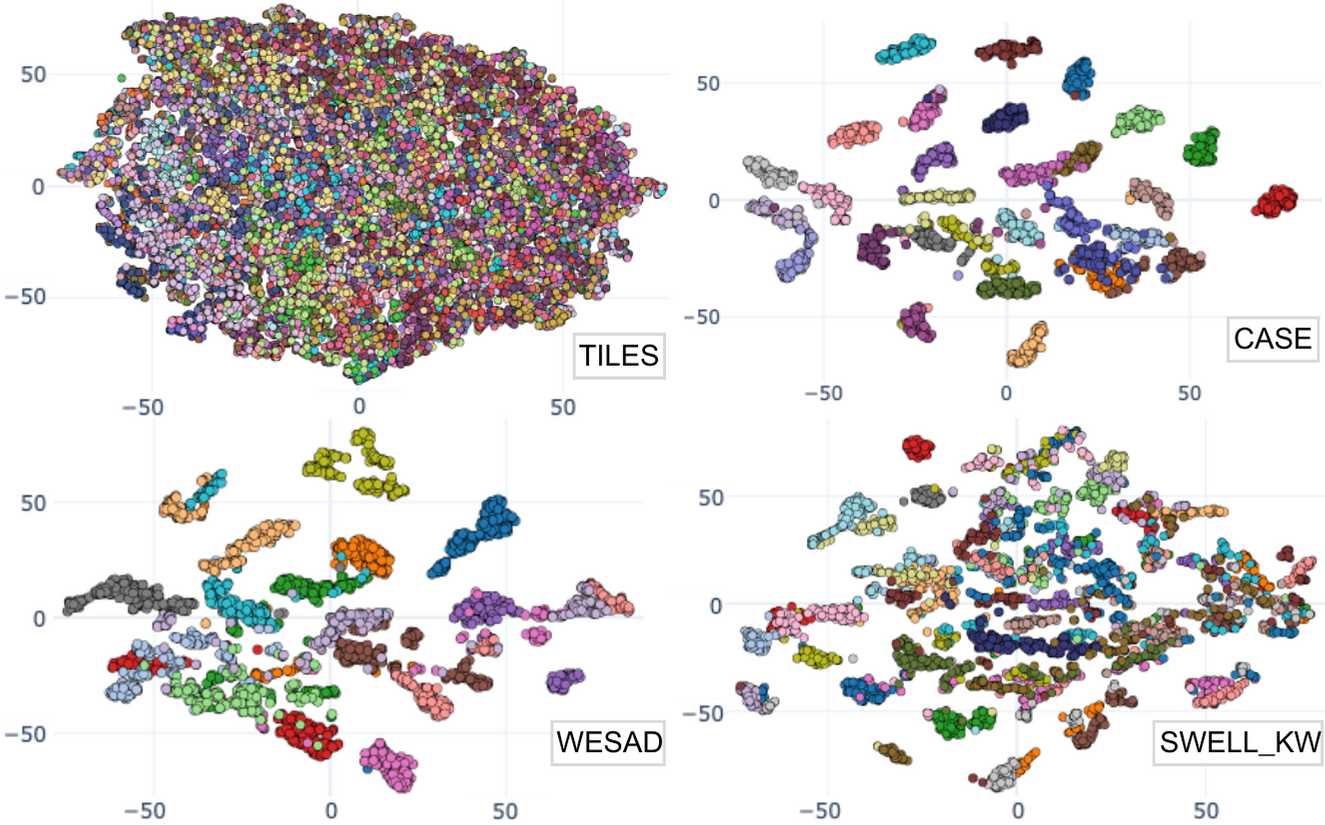}
    \caption{T-SNE visualizations of WildECG embeddings on the 2D space for TILES (downsampled to 10\%), CASE, WESAD and SWELL-KW datasets, colored by subject ID to reveal subject-specific bias.}
    \label{fig:tsne_participants}
\end{figure}

\begin{figure}
    \centering
    \includegraphics[scale=0.71]{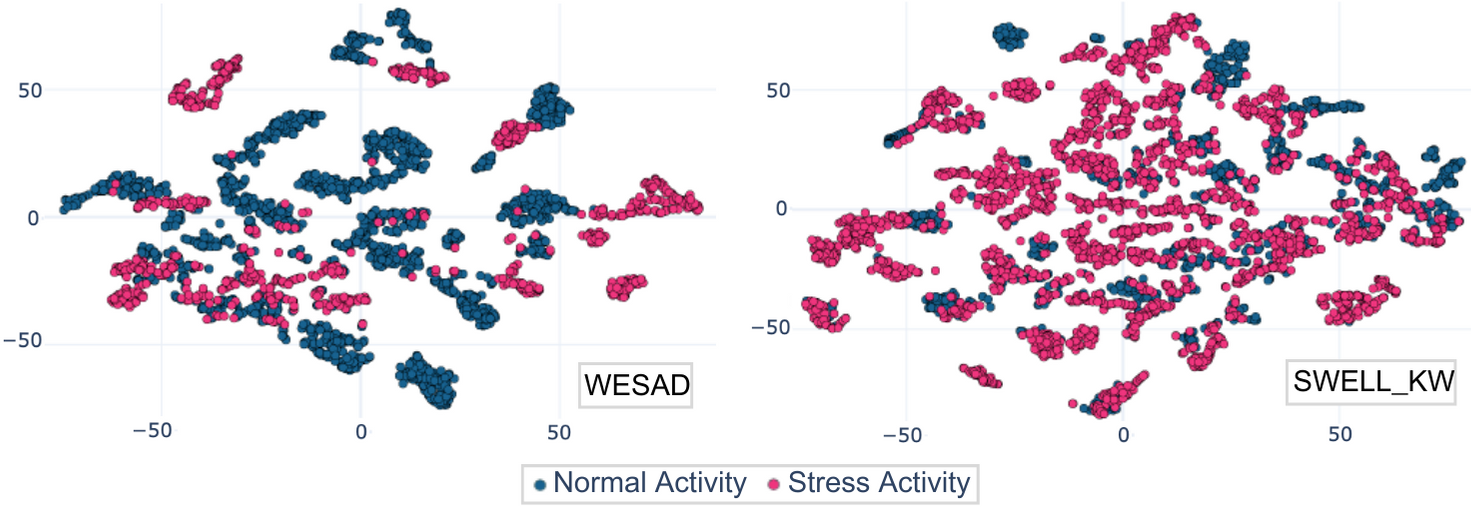}
    \caption{T-SNE visualizations of WildECG embeddings on the 2D space for WESAD and SWELL-KW  colored by type of activity.}
    \label{fig:tsne_classes}
\end{figure}

We first investigate whether WildECG incorporates subject-specific biases in its representations. To that end, we provide t-SNE visualizations colored by subject ID, in Figure~\ref{fig:tsne_participants}. As for the pre-training data, due to the number of samples TILES embeddings appear rather mingled, without visible clusters. However, the mean euclidean distance for intra-subject samples is found to be lower than the mean inter-subject distance, i.e., $7.02\pm 1.24$ vs. $10.55\pm2.59$, respectively; p-value $1.77\cdot10^{-27}$ for paired t-Student test. This indicates that samples related to the same subject are closer to each other rather than to samples from other subjects. On the other hand, for CASE, WESAD, and SWELL-KW, the embeddings form well-separated, participant-related clusters. This comes in spite of the subject-wise standardization and implies that the model indeed learns strong subject- or sensor-specific characteristics from the ECG. As for WESAD, each participant typically expands two clusters, grouped by heart rate value of the respective ECG. As shown in Figure~\ref{fig:tsne_classes}, these clusters indeed correspond to the \textit{stress} and \textit{no-stress} activities performed and in some cases form super-clusters across subjects.

\begin{figure}
    \centering
    \includegraphics[scale=0.61]{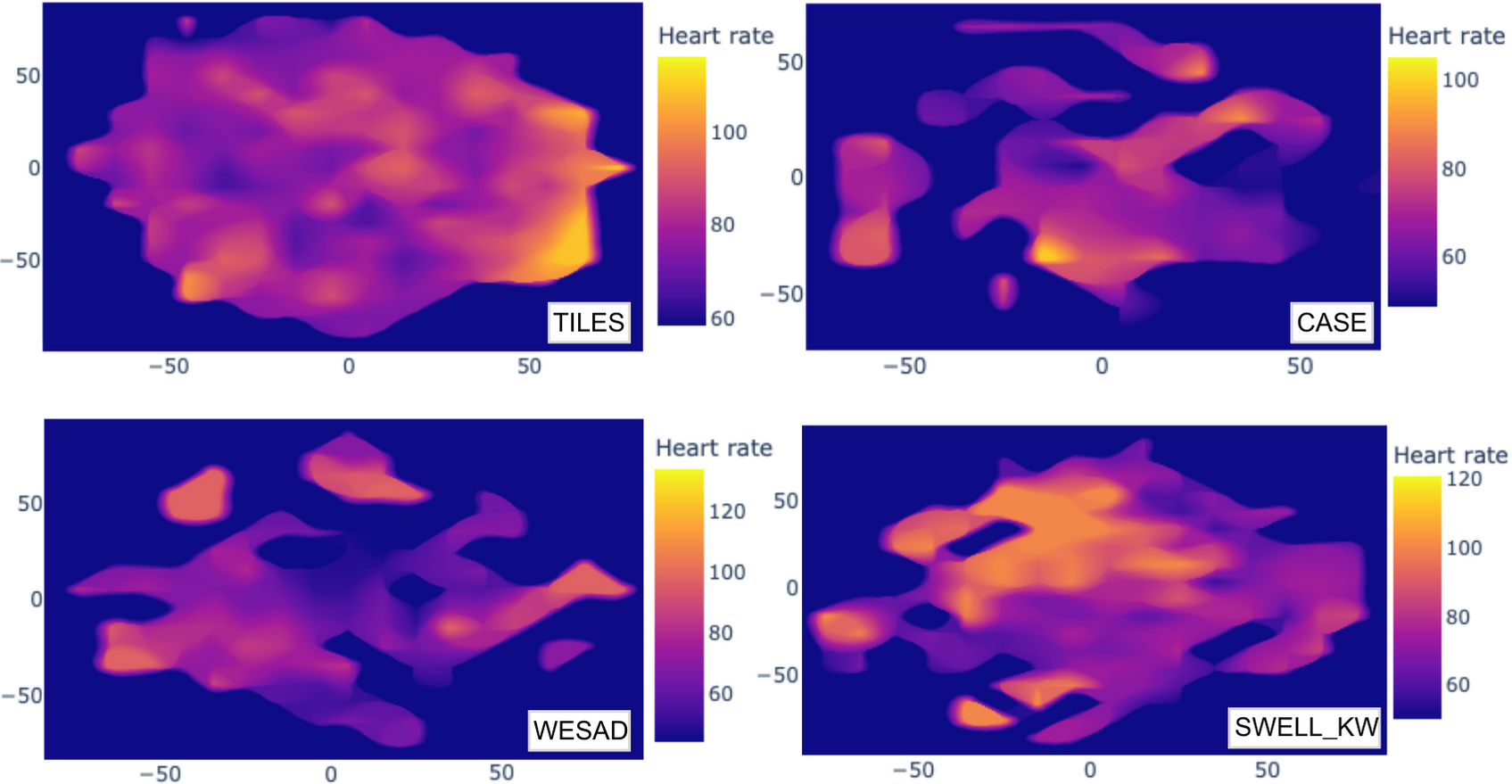}
    \caption{Distribution of heart rate values in 2D t-SNE space for WildECG embeddings of TILES  (downsampled), CASE, WESAD, and SWELL-KW samples. Brighter colors indicate higher heart rate.}
    \label{fig:hr_colored}
\end{figure}

Next, we focus on whether the WildECG representations retain cardiac information, in this case heart rate (HR). Ground truth HR values were obtained from the filtered 10-second ECG samples using the NeuroKit2~\cite{Makowski2021neurokit} library. The mean HR values were grouped into bins of 10 within the acceptable range of human HR~\cite{saganowski2022emotion}, i.e., from 40 to 210 bpm. In Figure~\ref{fig:hr_colored}, for TILES dataset we observe a left-to-right transition from low to higher HR values. On the contrary, CASE embeddings, which reflect strong subject biases (Figure~\ref{fig:tsne_participants}), do not present any generalized HR patterns. The transition across HR levels happens though within subject-specific clusters. Visualization for WESAD shows that the mixed-subjects' groups represent higher mean HR values, which are also associated with stress activity (Figure~\ref{fig:tsne_classes}). A similar pattern is also present in SWELL-KW, where samples from different subjects related to higher HR values and stress activity are close to each other.

\begin{figure}
    \centering
    \includegraphics[scale=0.7]{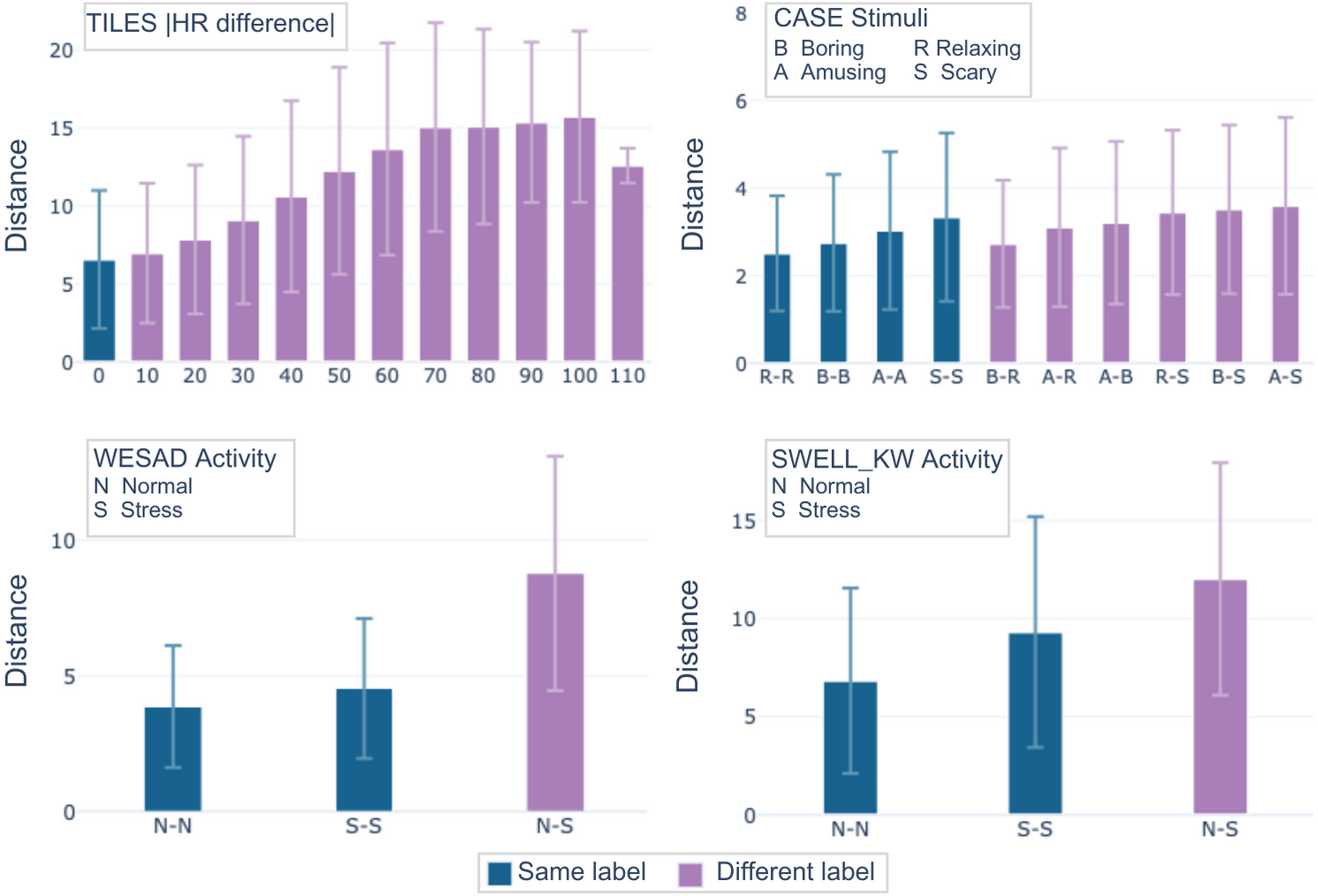}
    \caption{Average Euclidean distance of within-participant samples with respect to: absolute heart rate difference for TILES (downsampled), stimuli for CASE, and activity for WESAD and SWELL-KW. Colors indicate distances between samples with same or different labels. }
    \label{fig:distances_classes}
\end{figure}

To assess the model's capability to capture physiology that is characteristic of the context (activity or stimulus) we compute the Euclidean distance between high-dimensional embeddings of different context (Figure~\ref{fig:distances_classes}). For each subject, we determine the average distance between samples with the same label and the average distance between samples with different labels. The most notable difference between within-context and across-context samples can be observed for WESAD and SWELL-KW, where samples related to stressful activities are distant from other activities. As for CASE, the distinction between within-scary sample distance and distance to other-context samples is not significant, but remains noticeable. Further analysis of TILES data reveals that samples with similar HR values are situated closer to each other, while those with more significant differences in HR are positioned farther apart. This observation verifies that WildECG embeddings effectively preserve the cardiac information.

\subsection{Limitations and Challenges}

A robust and general-purpose representation model for the ECG is an important step towards expanding scientific research and clinical translation, notably for broad dissemination of smart and ubiquitous health applications. Equal importance however should be given to the limitations of these models, from the methodological aspects of evaluation, to fundamental questions regarding the applicability of physiological measures in estimating complex human conditions. Computational modeling of human behavior and physiology currently lacks standardized protocols and metrics that would ensure the reproducibility and validity of the obtained results. Consequently, our study also diverged from other comparable studies on how each dataset is set up for evaluation. However, we hope that it contributes a cohesive and comprehensive testbed for studies to evaluate their approaches.

An important characteristic of physiological signals that influences the evaluation protocol is the inter-subject variability, challenging the application of machine learning techniques like transfer learning in the field. Indeed, multiple studies report that machine learning models trained on a specific dataset rarely generalize to other datasets and settings~\cite{prajod2022generalizability,mishra2020evaluating}. Even within a specific dataset, subject bias could prevent evaluation on unseen subjects~\cite{sarkar2020self}. In our study, we demonstrate that pre-training with a large-scale ECG dataset in a self-supervised way can help alleviate these issues. However, our experiments and feature visualizations reveal that the learned features still reflect such biases, e.g., by forming subject-specific clusters. Hence, adopting specialized objectives to eliminate this bias is an important direction for future work.

Taking a step back, it is crucial to underscore the limitations of standalone measures like the ECG to solely estimate the range of human conditions. It is well known that emotional states are heavily influenced by the social and environmental context~\cite{barrett2019emotional, pikoulis2021leveraging}, in a way that a single-dimensional signal cannot reflect. We highlight that models like WildECG should be adopted in a holistic perspective that takes into consideration multiple views of human behavior and contextual information. For example, fusing information from multiple physiological and behavioral signal measures like electrodermal activity, speech~\cite{valstar2016avec}, and human activity~\cite{koldijk2016detecting} has provided better performance than using ECG alone on the same datasets. Incorporating WildECG in a multimodal sensing framework is another direction of future work to be pursued.

\section{Conclusion}

Ubiquitous sensing and monitoring are already transforming digital health and well-being with new, on-demand services. Hence there is an unmet need to address challenges related to the inference and analysis of the resulting rich and diverse human bio-behavioral states. In this study, we propose WildECG, a robust and versatile AI model for ECG representation learning. By utilizing a large, diverse corpus of biosignals collected in the wild, along with a state-of-the-art state-space network and pre-training algorithm, we demonstrate competitive performance on the tasks of estimating human affect, dimensional emotion, stress levels, as well as pathological markers. We further quantify the contributions of our design factors and verify model robustness in low-resource settings. The conducted qualitative analysis reveals that WildECG indeed incorporates explainable and tractable insights related to the ECG structure and features that could prove beneficial for researchers as well as clinicians.

\bibliographystyle{ACM-Reference-Format}
\bibliography{refs}

\section*{Appendix I \\ Related work on Self-Supervised Learning}
\label{related-ssl}

Self-supervised learning of biosignal representations, especially concerning the ECG modality, has recently shown rapid progress in health applications and behavioral analysis. In this section we include a comprehensive review of related studies that connect to our experimental methods and data.

Authors of \cite{sarkar2020self} used six signal transformations as pretext tasks to obtain ECG representation for downstream evaluation on affect recognition. They validated their obtained representation on four publicly available datasets. The results of this study in terms of model performance should be however interpreted with care since the introduced approach does not account for subject-specific biases in the training splits that could cause information leakage. In such an approach it is possible that two near-in-time samples will end up in training and test sets. The physiology across two near-in-time/consecutive samples (two 10-second long signal chunks) might be constant, especially in response to steady/constant/invariable stimuli.

In their analysis, the authors report that \%(1) the signal transformation parameters should be selected in such a way that the transformed signals are not too similar or too different from the original signal, because the model is unable to learn the representation, and that the most helpful transforms are scaling and time-warping.; (2) multiple proxy tasks build better representation than a single proxy task; (3) the most prominent proxy-tasks are scaling and time-warping; (4) representation learned on multiple datasets performs better than representation learned on a single dataset.

Among other studies, Kachuee et al. \cite{kachuee2018ecg} used ECG lead II signal of healthy and dysfunctional heart to recognize myocardial infarction. Their single-lead ECG SSL approach achieved results similar to a 12-lead ECG supervised approach. Dissanayake et al. \cite{dissanayake2022sigrep} focused on representation of low-frequency signals from wearable devices. They utilized SimCLR \cite{chen2020simple} and simplified Inception \cite{szegedy2015going} architectures in their pretext contrastive task. Their model performed competitively to the state-of-the-art results on publicly available datasets. Notably, their ablation studies showed that the model based on the representation is resistant to the loss of a large amount of training data, as well as to the loss of signal chunks.

Deldari et al.~\cite{deldari2022cocoa} proposed a cross-modal, contrastive, self-supervised learning approach to tackle several activity- and affect-related classification tasks. Using up to five modalities, their method outperformed several other approaches. According to their ablation analyses, as low as 10\% of the labels in the downstream task is adequate for the model to perform satisfactorily. Zhang et al. \cite{zhang2022self} used contrastive learning with the assumption that time- and frequency-based embeddings are located near each other in the time-frequency space. On the other hand, \cite{vazquez2022transformer} applied masking and attention to force the model to focus on relevant parts of the ECG signal. To build the signal representation, they trained on six public datasets and achieved competitive results on the AMIGOS~\cite{miranda2018amigos} dataset for binary arousal and valence classification.

\end{document}